\pdfoutput=1

\documentclass[11pt]{article}

\usepackage[final]{acl}
\usepackage{booktabs}

\usepackage{subcaption}  
\usepackage{enumitem}
\usepackage{bm}
\usepackage{times}
\usepackage{latexsym}

\usepackage[T1]{fontenc}

\usepackage{multirow}
\usepackage[table]{xcolor}
\usepackage{graphicx}  

\usepackage[utf8]{inputenc}

\usepackage{microtype}

\usepackage{inconsolata}

\usepackage{graphicx}
\usepackage{amsmath}

\usepackage{threeparttable}
\newcommand{\Same}{\ensuremath{\equiv}} 

\title{Paired by the Teacher: Turning Unpaired Data into High-Fidelity Pairs \\ for Low-Resource Text Generation}

\author{
\begin{tabular}{c}
Yen-Ju Lu, Thomas Thebaud, Laureano Moro-Velazquez,  \\ Najim Dehak, Jesus Villalba
\end{tabular}
\\
\begin{tabular}{c}
Center for Language and Speech Processing, Johns Hopkins University \\
\texttt{\{ylu125, tthebau1, laureano, ndehak3, jvillal7\}@jhu.edu}
\end{tabular}}

\begin{document}
\maketitle

\begin{abstract}
We present \emph{Paired by the Teacher} (PbT), a two-stage teacher–student pipeline that synthesizes accurate input–output pairs without human labels or parallel data. In many low-resource natural language generation (NLG) scenarios, practitioners may have only raw outputs, like highlights, recaps, or questions, or only raw inputs, such as articles, dialogues, or paragraphs, but seldom both. This mismatch forces small models to learn from very few examples or rely on costly, broad-scope synthetic examples produced by large LLMs. PbT addresses this by asking a teacher LLM to compress each unpaired example into a concise intermediate representation (IR), and training a student to reconstruct inputs from IRs. This enables outputs to be paired with student-generated inputs, yielding high-quality synthetic data. We evaluate PbT on five benchmarks—document summarization (XSum, CNNDM), dialogue summarization (SAMSum, DialogSum), and question generation (SQuAD)—as well as an unpaired setting on SwitchBoard (paired with DialogSum summaries). An 8B student trained only on PbT data outperforms models trained on 70 B teacher-generated corpora and other unsupervised baselines, coming within 1.2 ROUGE-L of human-annotated pairs and closing 82\% of the oracle gap at one-third the annotation cost of direct synthesis. Human evaluation on SwitchBoard further confirms that only PbT produces concise, faithful summaries aligned with the target style, highlighting its advantage of generating in-domain sources that avoid the mismatch, limiting direct synthesis.


\end{abstract}

\section{Introduction}


Many NLG tasks, such as summarization and question generation, lack readily available input–output pairs, often referred to as source–target pairs.
Creating these pairs requires experts to read each document or dialogue and write its summary or questions. This process is slow, costly, and must be repeated for every new domain. It also raises privacy concerns when conversations contain sensitive information.
To solve this problem, we introduce \emph{Paired by the Teacher} (PbT). Given only unpaired in-domain texts and a small set of example summaries or questions, PbT generates high-quality synthetic source–target pairs without any human pairing. This pipeline turns raw data into ready training material that matches the target style and speeds up adaptation across domains.


\begin{figure}[t]
  \centering
  \includegraphics[width=0.99\linewidth]{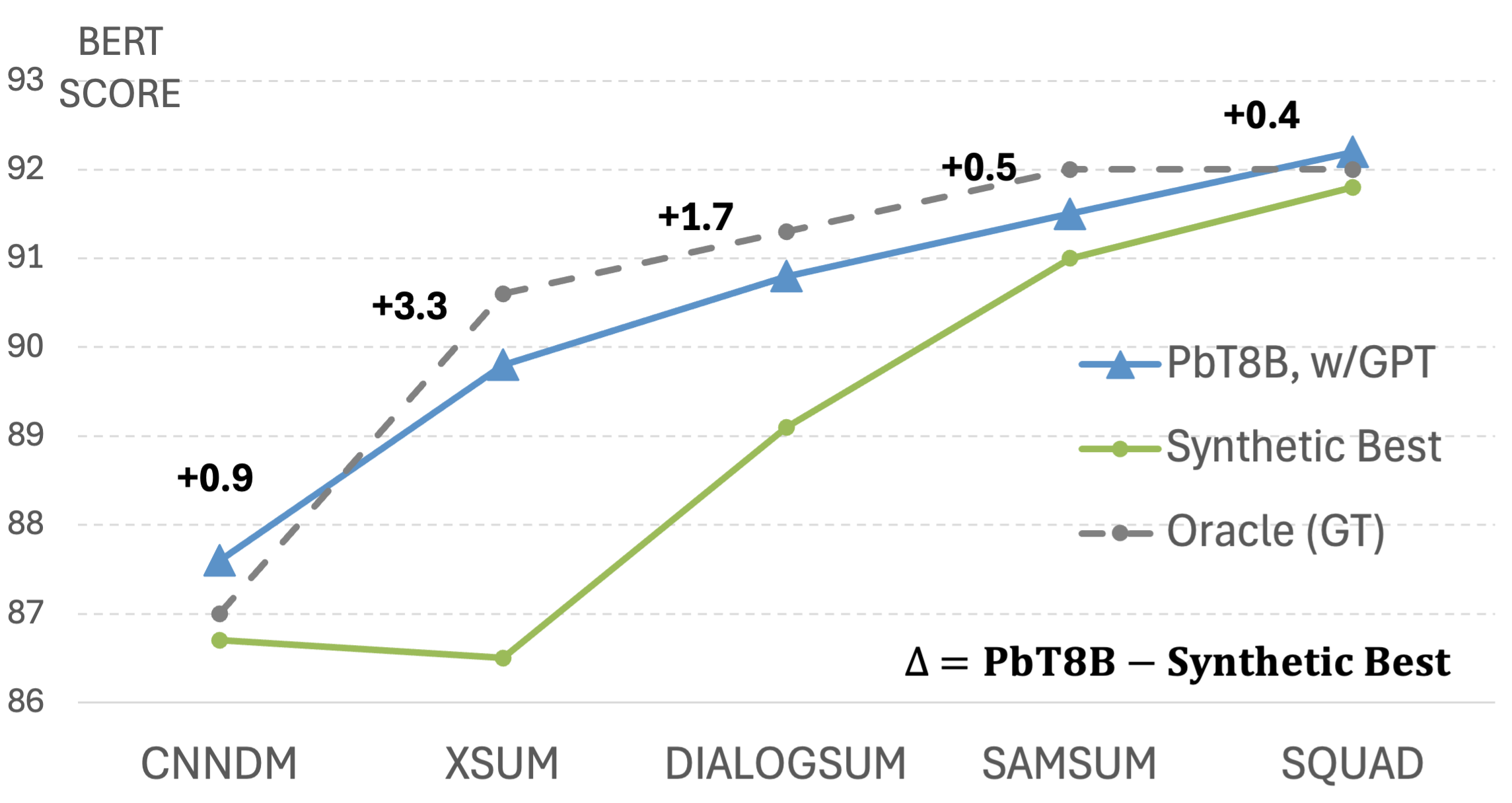}
  \caption{PbT consistently outperforms the strongest synthetic baselines from GPT and LLAMA70B (by up to +3.3 BERTScore points) and closes the gap to oracle (real) data across five summarization and QG tasks.}
  \label{fig:overview}
  \vspace{-3mm}
\end{figure}

Existing work shows that synthesis models can be trained using only unpaired data, such as back-translation~\citep{lample2018phrase}, GAN-based, or VAE-based summarization~\citep{wang2020learning, bravzinskas2020unsupervised}. More recent techniques leverage instruction-tuned teacher LLMs to generate training pairs, such as DiaSynth with GPT-4o \citep{suresh2025diasynth}, MixSumm with LLaMA-70B \citep{sahu2024mixsumm}, and ChatGPT self-chat corpora \citep{xu2023baize}. In few-shot scenarios, smaller teachers like GPT-3.5 or in-model synthesis adapters can also guide the pairing process \citep{mishra2023llm, lu2025mutual}.
However, these methods rely heavily on the teacher’s generation, refinement, and filtering abilities. This drives up inference costs and makes deployment in resource-constrained settings challenging, or requiring some ground-truth labeled data to ensure quality.

Paired by the Teacher (PbT) is a two-phase teacher–student pipeline that transforms unpaired inputs and outputs into high-quality source–target pairs without any manual annotation. In the first phase, a teacher model extracts a concise IR from each raw source example, and a smaller student model is fine-tuned via parameter-efficient methods to invert this mapping and reconstruct the source. 
In the second stage, the teacher model annotates IR for each unpaired target with the source-side IR example as in-context demonstrations. Each generated IR is then passed through the trained student model to synthesize an aligned source example. 
Because PbT asks the teacher to output only short IRs, the heavy text generation is offloaded to the student.  The teacher’s knowledge still guides the process, but its compute cost is a fraction of end-to-end generation.  At the same time, the student has been trained on real in-domain documents and dialogues. As a result, the synthetic sources it produces remain in-domain, unlike direct synthesis where mismatched sources hurt downstream performance. This in-domain alignment is the key reason PbT generalizes better at test time.
 
We evaluate PbT on five benchmarks—document summarization (XSum, CNNDM), dialogue summarization (SAMSum, DialogSum),  and question generation (SQuAD).
An 8B student trained only on PbT data outperforms models trained on 70B teacher-generated or self-generated data, and comes on average within 1.2 ROUGE-L points of human-annotated pairs, recovering 5.6 of the 6.8 ROUGE-L gap (82\% closure) compared with the best synthetic-only baseline.
We further build an unpaired setting on SwitchBoard dialogue-only dataset, paired with summaries from DialogSum and train the summarization model on purely cross-domain data.  
Human evaluation on SwitchBoard shows that only PbT adheres to the target summary lengths with concise, fluent, and faithful outputs. In contrast, all baselines produce overly long summaries due to a domain mismatch between the synthetic training data and the real test inputs. This highlights PbT’s core advantage: by generating sources through a student trained on real inputs, PbT keeps training data aligned with test-time conditions.

\paragraph{Major contributions:}
\begin{itemize}
  \item \textbf{Teacher–Student Pairing Pipeline.} PbT transforms unpaired sources and targets into high-fidelity training pairs, enabling a student model to learn from realistic source examples without any human annotations. Unlike direct synthesis, PbT’s student-generated sources remain in-domain, ensuring alignment between training and test distributions.
  \item \textbf{Robust Intermediate Representation.} We conduct the experiments with various IR formats and annotated by larger, same-sized, or smaller teachers, reducing dependence on model scale while maintaining data quality. Also, the cost for IR annotation is $3\times$ smaller than direct synthesis from the teacher model.
  \item \textbf{Intra-and Cross-Dataset Experiments:}  PbT not only surpasses teacher/student-only baselines across all five benchmarks in automatic metrics and model-based evaluation, but also yields a functional summarizer on the label-free SwitchBoard corpus by borrowing DialogSum summaries—adhering to their style and outperforming baseline systems in both human and GPT evaluation.

\end{itemize}

\begin{figure}[t]
  \centering
  \includegraphics[width=0.99\linewidth]{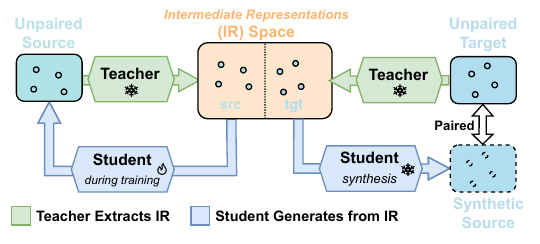}
  \caption{High-level overview of Paired-by-the-Teacher (PbT). The teacher LLM extracts intermediate representations (IR) from both unpaired sources (left) and unpaired targets (right). A student model is then trained to invert IR$\rightarrow$source (blue loop) and to synthesize new sources from IR (blue arrow), producing synthetic pairs $(\tilde x,y)$ for downstream NLG fine-tuning.}
  \label{fig:overview}
\end{figure}

\begin{figure*}[th]
    \centering
    \begin{subfigure}{0.49\linewidth}
        \centering
        \includegraphics[width=\linewidth]{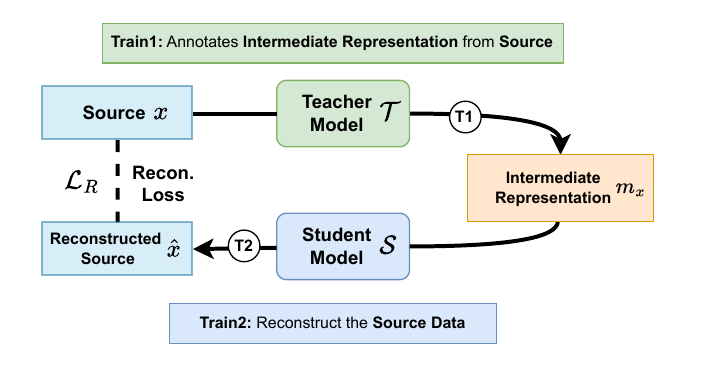}
        \caption{\textbf{Source-side IR Learning (Student Reconstruction).} 
       }
        \label{fig:pbt_train}
    \end{subfigure}
    \begin{subfigure}{0.48\linewidth}
        \centering
        \includegraphics[width=\linewidth]{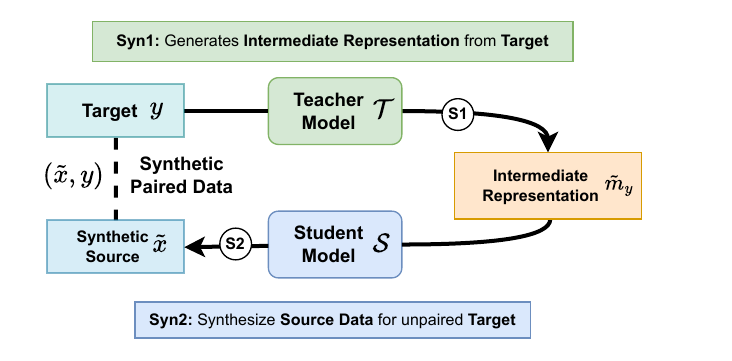}
        \caption{\textbf{Target-side IR Annotation (Synthetic Pair Generation).}
       }
        \label{fig:pbt_syn}
    \end{subfigure}
    \caption{Overview of the \emph{Paired by the Teacher} (PbT) pipeline. 
    (a) The teacher model $\mathcal{T}$ extracts an intermediate representation $m_x$ from each source $x$, and the student $\mathcal{S}$ learns to reconstruct $x$ from $m_x$. 
    (b) The teacher annotates each target $y$ with a plausible IR $\tilde m_y$, which the student expands into a synthetic source $\tilde x$, forming synthetic pairs $(\tilde x, y)$ for downstream training without human-written labels.
    }
    \label{fig:PbT}
    \vspace{-3mm}
\end{figure*}

\section{Related Work}
\paragraph{Unpaired and Low-Resource NLG with Model Selection}
Before the emergence of LLMs, unsupervised text generation used cycle objectives or auto-encoding on smaller models. Back-translation let dual translators learn without parallel data \citep{lample2018phrase}. Similar cycle or reconstruction ideas powered unsupervised summarization, sentence compression, and opinion summarization \citep{wang2020learning,baziotis2019seq,bravzinskas2020unsupervised}, as well as large-scale pseudo-pre-training from news leads \citep{yang2020ted}. Cycle consistency was later applied to data-to-text generation \citep{guo2020cyclegt}. Because no gold validation exists, these studies chose checkpoints with round-trip BLEU, reconstruction loss, or discriminator scores, which correlate well with true quality. We follow the same principle, using IR-to-document reconstruction to guide model selection for LLM-scale training.

\paragraph{LLM-Driven Pseudo-Labeling and In-Context NLG}  
With instruction-tuned LLMs, generating synthetic corpora for low-resource tasks has become widespread. MixSumm~\citep{sahu2024mixsumm} and PPSL~\citep{sahu2025guide} prompt LLaMA-3-70B to synthesize documents and pseudo-summaries, then fine-tune a 100–300 M summarizer to match fully-supervised ROUGE with only 5 \% gold pairs~\citep{sahu2025guide}. 
\citet{mishra2023llm} use GPT-3.5 to pseudo-label customer-service chats; with just 10 \% human data, their BERT-based model recovers 95 \% of full-data performance. 
MRDS \citep{lu2025mutual} further intertwines dialogue synthesis and summarization LoRA adapters inside a single LLM with pseudo-pair generation. 
In parallel, self-chat methods like Baize \citep{xu2023baize} collect $\sim$100 k ChatGPT-generated dialogues to LoRA-tune LLaMA, approaching proprietary chatbot quality, and task-oriented work \citep{surana2025reviews} turns product reviews into synthetic recommendation dialogues for ReDial~\citep{li2018towards} and INSPIRED~\citep{hayati2020inspired} benchmarks.
Finally, zero-shot and in-context prompting with GPT-3 can produce news summaries preferred by humans over gold references~\citep{goyal2022news}, though such methods lack a lightweight deployable model, motivating hybrid approaches that combine in-context learning with pseudo-fine-tuning.

\section{Methodology}

\emph{Paired by the Teacher} (PbT) is a unified teacher–student pipeline that converts unpaired inputs and outputs into high-fidelity source–target pairs without any human labeling. 

\paragraph{PbT Overview}
Figure~\ref{fig:overview} provides a high-level illustration of the PbT pipeline. The process unfolds in three phases. 
In the {Source IR Induction} phase, the teacher LLM compresses each unpaired source into a concise intermediate representation (IR), and the student model is trained to reconstruct the original input from this IR. 
In the {Target IR Annotation} phase, the teacher annotates each unpaired target with a plausible IR, which the trained student then expands into a synthetic source, forming paired examples $(\tilde{x},y)$. 
Finally, in the \emph{Downstream Fine-tuning} phase, a summarization or question-generation model is fine-tuned on these synthetic pairs. 
By shifting heavy generation to the student while keeping the teacher’s guidance at the IR level, PbT achieves both efficiency and alignment  (see Appendix~\ref{sec:prompt} for concrete IR examples).

\subsection{Source-side IR Learning}

Our pipeline begins by teaching the student model to reconstruct in-domain source data from the IR. 
We first use a large teacher model $\mathcal{T}$ to extract a structured IR from each raw source example, and then fine-tune a smaller student model $\mathcal{S}$ to invert that mapping.

\paragraph{IR Extraction (Teacher)} 
As illustrated in Figure~\ref{fig:PbT}a, for every unpaired source instance $x$ (dialogue, article, paragraph, ...), the teacher $\mathcal{T}$ produces a concise IR $m_x$ (e.g., a topical outline, chain-of-thought, or related facts). This yields a semi-supervised dataset of $(x_i, m_{x_i})$ pairs:
\[
\{(x_i, m_{x_i})\}_{i=1}^N . \tag{1}
\]

The form of $m_{x_i}$ is task-dependent. For dialogue and document summarization, the teacher is instructed to split the input into paragraph groups and write a topic sentence with key entities. For question generation, the teacher extracts the answer-containing sentence and writes supporting facts as bullets. These task-specific IRs provide the information needed to reconstruct the source $x_i$ (see Table~\ref{tab:ir_source}).

\paragraph{IR-to-Source Induction (Student)} 
We then fine-tune the student model $\mathcal{S}$ with parameter-efficient methods (PEFT) on these source-side annotations to learn the inverse mapping
\[
\mathcal{S}_{m\rightarrow x}: m \mapsto x . \tag{2}
\]
By optimizing on $(x_i, m_{x_i})$, the student internalizes the source domain’s structural patterns and can expand IRs into realistic source texts $\hat{x} = \mathcal{S}_\theta(m)$. The model is trained with the reconstruction loss
\[
\theta = \arg\min_\theta \sum_{i=1}^N \mathcal{L}_R(\mathcal{S}_\theta(m_{x_i}), x_i), \tag{3}
\]
where $\mathcal{L}_R$ is the cross-entropy loss. These learned synthesis skills later allow $\mathcal{S}$ to generate candidate sources for unpaired targets.

\subsection{Target-side IR Annotation}

\paragraph{IR Annotation (Teacher)}  
As shown in Figure~\ref{fig:pbt_syn}, we prompt the teacher LLM $\mathcal{T}$ to annotate each unpaired target example $y$ with a plausible IR $\tilde{m}_y$. To ensure consistent formatting, $\mathcal{T}$ is provided with a few in-context examples of $(x,m_x)$ pairs from the source-side training phase. This yields a set of target annotations:
\begin{equation}
\{(y_j, \tilde{m}_{y_j})\}_{j=1}^{N_y}\,.
\end{equation}

Because the teacher does not observe the corresponding source $x_j$, the generated $\tilde{m}_{y_j}$ are not aligned with the actual sources. Such explicit alignment is unnecessary, as the student generates synthetic sources directly from these IRs.

\paragraph{IR-to-Source Generation (Student)}  
Once the student model $\mathcal{S}_{m\to x}$ has learned to expand IRs into full source texts, we apply it to each target-side representation $\tilde m_{y_j}$. 
The student then produces a synthetic source $\tilde x_j$ for each target $y_j$, resulting in a synthetic paired corpus
\begin{equation}
\{(\tilde x_j,\,y_j)\}_{j=1}^{N_y}\,,
\end{equation}
where every target is matched with a candidate source generated solely from its corresponding IR.

\subsection{Target Task Training}
Finally, we fine-tune a downstream student model $\mathcal{S}_{x\to y}$ on the synthetic pairs $\{(\tilde x_j,\,y_j)\}_{j=1}^{N_y}$. The model learns to map synthetic sources to targets:
\begin{equation}
\mathcal{S}_{x\to y}\;:\;\tilde{x}\;\mapsto\;y .
\end{equation}

The training objective is
\begin{equation}
\phi = \arg\min_\phi \sum_{j=1}^{N_y} \mathcal L_S\bigl(\mathcal{S}_\phi(\tilde x_j),\,y_j\bigr),
\label{eq:S_x2y}
\end{equation}
yielding predictions $\hat y = \mathcal{S}_\phi(\tilde x)$.

PbT mimics the effect of having true training pairs for two reasons. 
First, it generates only synthetic source inputs $\tilde x$ and pairs them with authentic targets $y$, avoiding propagation of synthesis artifacts into outputs. 
Second, by training the student on real source data, PbT ensures that synthetic sources remain in-domain. 
As a result, at test time, PbT-generated pairs better match the distribution of real inputs, unlike direct synthesis, which suffers from mismatched sources.

\begin{table*}[ht]
\centering

\begin{threeparttable}
\caption{\textbf{XSum (200-shot unpaired documents)}: Comparison of methods on small-source settings.}

\vspace{-2mm}
\label{tab:xsum}
\begin{tabular}{@{}lccc|cccc|c@{}}
\toprule
\multirow{2}{*}{Synthesis model}
  & \multicolumn{3}{c}{Ref-based}
  & \multicolumn{5}{c}{G-Eval}
   \\
\cmidrule(lr){2-4} \cmidrule(l){5-9}
  & R-2 & R-L & BERTScore
  & Coh. & Con. & Rel. & Flu. & Ave.
 \\ 
\midrule
Zero shot           &  5.3  & 12.5  & 86.2  & \multicolumn{5}{c}{N/A$^\dagger$}      \\
ICL (summary-only)  &  7.9  & 18.2  & 87.6  &  \multicolumn{5}{c}{N/A$^\dagger$}      \\
\midrule
\multicolumn{9}{l}{\textbf{Unpaired Sources — Direct synthesis} (Summ→(\textit{Teacher/Student})→Document)} \\
\midrule
LLAMA8B$_{\text{direct}}$ ($\mathcal{S}$)   
  &  6.5  & 18.4  & 87.0  
  & 2.29  & 3.83  & 2.09 & 2.95
  & 2.79  \\
LLAMA70B$_{\text{direct}}$ ($\mathcal{T}$) 
  &  5.0  & 16.5  & 86.5  
  & 2.33  & 4.03  & 2.09 & 2.94
  & 2.84  \\
GPT$_{\text{direct}}$ ($\mathcal{T}$)       
  &  5.0  & 16.4  & 86.2  
  & 2.22  & 3.78  & 2.09 & 2.91
  & 2.75  \\
GPT$_{\text{direct, BoN-5}}$ ($\mathcal{T}$)       
  &  5.4  & 16.9  & 86.5  
  & 2.38  & 3.95  & 2.25 & 2.92
  & 2.88  \\
\midrule
\multicolumn{9}{l}{\textbf{Unpaired Sources — PbT synthesis} (Summ→\textit{Teacher}→{IR}→\textit{Student}→Document)} \\
\midrule
PbT$_\text{8B, w/ llama70B}$   
  & 16.0  & 29.3  & 89.7     
  & 2.95  & {4.14}  & 2.79 & \textbf{3.00}
  & {3.22}  \\
PbT$_\text{8B, w/ GPT}$   
  & \textbf{16.1}  & \textbf{29.4}  & \textbf{89.8}  
  & {2.99}  & 4.08  & {2.85} & 2.98
  & {3.22}  \\
PbT$_\text{8B, w/ GPT, BoN-5}$   
  & 15.4  & 28.6  & 89.6
      & \textbf{3.18}  & \textbf{4.43}  & \textbf{3.10} & \textbf{3.00}
  & \textbf{3.43}  \\
\midrule
\multicolumn{9}{l}{\textbf{Oracle Paired}} \\
\midrule
Real Data           
  & 18.7  & 32.6  & 90.6  
  & 2.74  & 3.71  & 2.59 & \textbf{3.00}
  & 3.01  \\
\bottomrule
\end{tabular}
\begin{tablenotes}[para,flushleft]
\footnotesize
\item $\dagger$ N/A indicates that G-Eval results are omitted for pretrained LLM outputs due to known evaluation bias.
\end{tablenotes}
\end{threeparttable}
\vspace{-2mm}
\end{table*}


\section{Experiments}
\paragraph{Intra-Datasets Experiment} We evaluate PbT on three tasks with five benchmarks: dialogue-summarization (DialogSum~\citep{chen2021dialogsum} and SAMSum~\citep{gliwa2019samsum}), document summarization (XSum~\citep{narayan2018don} and CNNDM~\citep{hermann2015teaching}), and question generation (SQuAD~\cite{rajpurkar2016squad}). 
For each dataset, we perform intra-dataset experiments by partitioning the data into "source-only” (dialogues, documents, and paragraphs) and "target-only” (summaries and questions) subsets, training the model without any paired examples. 

\paragraph{Cross-Datasets Experiment} 
Additionally, we assess the cross-domain transfer ability of PbT by training it on SwitchBoard~\citep{godfrey1992switchboard}, a conversation-style dialogue-only dataset, containing 1155 dialogues without any available summary labels.
We train the PbT model with the dialogues from SwitchBoard and summaries from DialogSum, aiming to train a summarization model that can summarize the dialogues from SwitchBoard into DialogSum-style summaries.

\begin{table*}[ht]
\centering
\begin{threeparttable}
\caption{\textbf{CNNDM (200-shot unpaired Documents)}: Comparison of methods on small-source settings.}
\vspace{-2mm}
\label{tab:cnndm_small}
\begin{tabular}{@{}lccc|cccc|c@{}}
\toprule
\multirow{2}{*}{Synthesis model}
  & \multicolumn{3}{c}{Ref-based}
  & \multicolumn{5}{c}{G-Eval}
   \\
\cmidrule(lr){2-4} \cmidrule(l){5-9}
  & R-2 & R-L & BERTScore
  & Coh. & Con. & Rel. & Flu. & Ave.
 \\ 
\midrule
Zero shot           &  13.1  & 21.0  & 86.8  
  & \multicolumn{5}{c}{N/A$^\dagger$} \\
ICL (summary-only)  &  15.2  & 24.6  & 87.5  & \multicolumn{5}{c}{N/A$^\dagger$} \\ 
\midrule
\multicolumn{9}{l}{\textbf{Unpaired Sources}} \\
\midrule
GPT$_\text{direct}$ ($\mathcal{T}$)       
  &  16.0  & 22.7  & 86.4  
  &  4.51  & 4.92  & 4.47  & 2.94  & 4.21 \\
LLAMA8B$_\text{direct}$ ($\mathcal{S}$)   
  &  16.6  & 23.9  & 86.7  
  &  4.47  & \textbf{4.95}  & 4.29  & 2.94  & 4.16 \\
PbT$_\text{8B, w/ GPT}$   
  & \textbf{17.5}  & \textbf{27.3}  & \textbf{87.6}  
  &  \textbf{4.61}  & 4.91  & \textbf{4.53}  & \textbf{2.98}  & \textbf{4.26} \\
\midrule
Oracle Paired           
  &  16.8  & 25.7  & 87.0  
  &  4.41  & 4.90  & 4.21  & 2.93  & 4.11 \\
\bottomrule
\end{tabular}
\begin{tablenotes}[para,flushleft]
\footnotesize
\item $\dagger$ N/A indicates that G-Eval results are omitted for pretrained LLM outputs due to known evaluation bias.
\end{tablenotes}
\end{threeparttable}
\vspace{-2mm}
\end{table*}

\subsection{Alternative Methods}

We compare our PbT method with several unpaired summarization alternative approaches, divided into two categories: methods using the pre-trained Llama3 model without fine-tuning and methods fine-tuned on real or synthetic data.
\paragraph{Pre-trained Methods}
\begin{itemize} [label=-, topsep=0pt, itemsep=1pt, parsep=0pt, partopsep=0pt]
\item \textbf{Zero-shot:} Zero-shot summarization performance of LLAMA8B. 
\item \textbf{ICL:} Summarization performance of LLAMA8B using in-context learning with $k=5$ summaries examples.
\end{itemize}
\paragraph{Fine-tuned Methods}
\begin{itemize} [label=-, topsep=0pt, itemsep=1pt, parsep=0pt, partopsep=0pt]
\item \textbf{Direct Synthesis}: Using a fixed teacher or student model to generate source from the summaries with in-context source examples. 
Then we fine-tune the student summarization model from those synthetic source-target pairs.
\item \textbf{Best-of-5 Filtering (BoN-5)}: Applied to both baselines and PbT, this strategy generates five candidate pairs and filter as in Genie~\citep{yehudaiachieving}. GPT-4o-mini is first used to filter inconsistent document–summary pairs, and a QRM-Gemma-2-27B ranker~\citep{dorka2024quantile} selects the top candidate.

\item \textbf{Oracle Paired}: Fine-tuning the student summarization model with the oracle summaries for the source shots used in other experiments. 

\item \textbf{DiaSynth}: Fine-tuning the student summarization model on synthetic data from~\citep{suresh2025diasynth}, which leverages LLM CoT reasoning. The amount of synthetic training data is aligned with the other experiments.

\end{itemize}

\subsection{Implementation Details}
\paragraph{Teacher/Student Selection} 
Our primary experiments use Llama3.1-8B-Instruct \citep{dubey2024llama} as the student model (denoted LLAMA8B). We train LoRA adapters of rank 16 and $\alpha=32$ on top of the frozen base weights. To generate IRs, we rely on two large teachers: the Llama3.3-70B-Instruct model (LLAMA70B) and GPT-4o mini (GPT), and then fine-tune LLAMA8B on the synthetic data they produce. 

\paragraph{Training Settings (Summarization, QG, and Source-synthesis).}
Both the summarization / question-generation models and the IR $\!\rightarrow$ source synthesizer are trained with the same optimization recipe from~\citep{lu2025mutual}.
During synthesis training, the student sees only the IR without summaries, forcing them to reconstruct the source from the IR alone.
For the proposed methods in the main experiments, we run the synthesis and training process three times and compute the average.
The hyperparameters and prompts for IR extraction, source generation, and downstream training are listed in Appendix~\ref{sec:hyp}~\ref{sec:prompt}.

\begin{table*}[ht]
\centering
\begin{threeparttable}
\caption{\textbf{DialogSum (124/623-shot unpaired dialogues)}: Comparison of methods on small- and medium-source settings.}

\vspace{-2mm}
\label{tab:dialogsum}
\begin{tabular}{@{}lccc|ccc@{}}
\toprule
\multirow{2}{*}{Synthesis model}
  & \multicolumn{3}{c}{Small (124-shot)}
  & \multicolumn{3}{c}{Medium (623-shot)} \\
\cmidrule(lr){2-4} \cmidrule(l){5-7}
  & R-2 & R-L & BERTScore
  & R-2 & R-L & BERTScore \\
\midrule
Zero shot                   & 9.0   & 19.6 & 87.3   & \Same  & \Same  & \Same   \\
ICL (summary-only)          & 13.8    & 28.7   & 89.9      & \Same  & \Same  & \Same   \\ 
\midrule
\multicolumn{7}{l}{\textbf{Unpaired Sources — Direct synthesis} (Summ→(\textit{Teacher/Student})→Dialogue)} \\
\midrule
DiaSynth & 11.9 & 25.0 & 88.8  & \Same  & \Same  & \Same  \\ 
GPT$_\text{direct}$ ($\mathcal{T}$)          & 14.7  & 27.7 & 89.1   & 15.4  & 29.2 & 89.5   \\
LLAMA70B$_\text{direct}$ ($\mathcal{T}$)    & 15.2    & 28.7   & 89.1     & 15.1    & 28.9   & 89.3     \\
LLAMA8B$_\text{direct}$ ($\mathcal{S}$)      & 14.4  & 29.2 & 89.4   & 13.7  & 27.3 & 89.0   \\
\midrule
\multicolumn{7}{l}{\textbf{Unpaired Sources — PbT synthesis} (Summ→\textit{Teacher}→IR→\textit{Student}→Dialogue)} \\
\midrule
PbT$_\text{8B, w/ GPT}$  & \textbf{17.7} & \textbf{35.0} & \textbf{90.8}
                            & \textbf{18.4}  & \textbf{35.3} & \textbf{91.2}   \\
PbT$_\text{8B, w/ llama70B}$   & \textbf{17.8}    & \textbf{34.1}   & \textbf{90.4}     & \textbf{18.7} & \textbf{35.1} & \textbf{90.9} \\ 
\midrule
\multicolumn{7}{l}{\textbf{Oracle Paired}} \\
\midrule
Real Data               & 19.6  & 36.9 & 91.3   & 20.6  & 37.4 & 91.3   \\
\bottomrule
\end{tabular}

\begin{tablenotes}[para,flushleft]
\footnotesize
\item \Same\ indicates the result is the same as in the Small setting.
\end{tablenotes}
\end{threeparttable}

\vspace{-2mm}
\end{table*}

\section{Results}

\subsection{Unpaired Document Summarization}
For document summarization, we conduct experiments at two different scales—200-shot and 1,000-shot source training—on the XSum and CNNDM datasets. For validation and testing, we randomly sample 500 and 1,000 examples, respectively, from each split. In addition to ROUGE and BERTScore, we also report G-Eval~\citep{liu2023g}, a rigorously designed model-based evaluation framework that uses fixed prompts and the GPT-4 model, and has shown high correlation with human judgments. 

\paragraph{XSum}
As shown in Table~\ref{tab:xsum}, XSum is challenging dataset, as it requires generating a one-line summary from a long news article. Both zero-shot and ICL experiments yield ROUGE-2 scores below 10\%, and direct synthesis performs even worse than ICL (7.9 vs. 5.0 and 6.5 R-2), falling far short of the oracle paired-data baseline (18.7 R-2). In contrast, the proposed PbT approach achieves a ROUGE-2 score of 16.1, closing more than 80\% of the gap to the oracle performance. Incorporating the Best-of-5 (BoN-5) filtering strategy does not further improve ROUGE-based scores, suggesting limited gains under reference-based evaluation. Results for the medium-scale XSum setting are provided in Appendix~\ref{sec:medium}.

\paragraph{XSum G-Eval}
Table~\ref{tab:xsum} also presents G-Eval metrics on XSum, including Coherence (Coh.), Consistency (Con.), Relevance (Rel.), and Fluency (Flu.). We compute over the first 200 test examples across 20 runs. We find that the direct-synthesis baselines fall short in coherence and relevance. In contrast, PbT outperforms the real paired data on three metrics, achieving an overall average of 3.22 compared to 3.01. With BoN-5 filtering, although no further improvements are observed in Ref-based scores, PbT achieves a new best overall G-Eval score of 3.43, showing consistent gains in coherence, consistency, and relevance. This indicates that BoN-5 effectively refines the quality of synthetic data under unpaired scenarios when evaluated by human-aligned metrics.

\paragraph{CNNDM}
We also conduct experiments on CNNDM, a dataset with higher risk of pretraining contamination~\citep{jiang2024investigating}. As a result, the gap between pretrained models and oracle paired data is relatively small—only 2--4\% in ROUGE-L (21.0, 24.6 vs. 25.7)—compared to the much larger gap on XSum (12.5, 18.2 vs. 32.6). In this setting, direct synthesis offers only modest ROUGE-2 gains, whereas PbT outperforms both ICL and oracle paired data. Moreover, G-Eval shows that PbT achieves the best overall average of 4.26 (vs. 4.11 for oracle), with notable improvements in relevance (4.53 vs. 4.21) and coherence (4.61 vs. 4.51), highlighting that PbT yields more human-aligned quality gains even when reference-based improvements are limited.

\begin{table}[ht]
\centering
\caption{\textbf{SAMSum (147-shot unpaired dilaogues)}: Comparison of methods on the small‑size source data.}
\vspace{-2mm}
\label{tab:samsum_small}
\begin{tabular}{@{}lccc@{}}
\toprule
Synthesis model            & R-2   & R-L   & B-S \\
\midrule
Zero‑shot                  & 12.9  & 25.1  & 88.3 \\
ICL (summary only)         & 18.5  & 33.8  & 90.4 \\
\midrule
\multicolumn{4}{l}{\textbf{Unpaired Sources}} \\
\midrule
DiaSynth                  & 16.0  & 29.8  & 89.5 \\
GPT$_\text{direct}$             & 23.6  & 37.7  & 90.9 \\
LLAMA8B$_\text{direct}$           & 22.6  & 37.2  & 91.0 \\

PbT$_\text{8B, w/ GPT}$       & \textbf{24.6} & \textbf{40.3} & \textbf{91.5} \\
\midrule
Oracle Paired              & 27.9  & 43.7  & 92.0 \\
\bottomrule
\end{tabular}
\vspace{-2mm}
\end{table}

\subsection{Unpaired Dialogue Summarization}
We evaluate PbT on DialogSum and SAMSum under two resource settings: 
small (1\% of dialogues, 124/147-shot) and medium (5\%, 623/736-shot) isolated source data. 
For fair comparison across datasets, we fix 1,000 unpaired targets in all experiments.
The evaluation includes ROUGE-2 (R-2), ROUGE-L (R-L) ~\citep{lin2004rouge}, and BERTScore ~\citep{bert-score}.

\begin{table*}[ht]
\centering
\caption{\textbf{SQuAD (200-shot unpaired paragraphs)}: Comparison of question-generation pipelines. Metrics on the left (ROUGE-2, ROUGE-L, BERTScore, BLEU, METEOR) measure surface similarity to the reference question; columns on the right evaluate the usefulness of each generated question by running a QA model on the original passage (Exact-Match / F1).}
\vspace{-2mm}
\label{tab:SQuAD}
\begin{tabular}{@{}lccccccc@{}}
\toprule
Synthesis model  & R-2 & R-L & BERTScore
  & BLEU   & METEOR  & QA EM. & QA F1. \\
\midrule
Zero shot           &  25.1 & 41.6 & 91.2 & 12.8 & 41.6 & \textbf{53.0} & \textbf{72.2}      \\
ICL (summary-only)  & 24.1 & 40.8 & 91.0 & 13.0 & 39.4 & 47.3 & 65.3      \\
\midrule
\multicolumn{7}{l}{\textbf{Unpaired Sources}} \\
\midrule
GPT$_{\text{direct}}$ ($\mathcal{T}$)       
                     &  27.8  & 45.5 & 91.8 & 18.8 & 39.3 &47.7 &66.8  \\
LLAMA8B$_{\text{direct}}$ ($\mathcal{S}$)   
                     &  27.0 & 44.5 & 91.6 &  17.7 & 37.1 & 46.7 & 64.2\\
PbT$_\text{8B, w/ GPT}$   & \textbf{30.3} & \textbf{48.0} & \textbf{92.2} & \textbf{20.3} & \textbf{41.1} & \textbf{52.3} & \textbf{71.1} \\ 
\midrule
Oracle Paired           & 29.5 & 47.2 & 92.0 & 19.4 & 39.7 & 51.4 & 70.8  \\ 
\bottomrule
\end{tabular}
\vspace{-2mm}
\end{table*}

\begin{table}[t]
\caption{\textbf{SwitchBoard → DialogSum (1155 unpaired dialogues)}: Cross-dataset transfer results for dialogue summarization.}

\vspace{-2mm}
\label{tab:swb}
\centering
\begin{tabular}{@{}lcc|cc|c@{}}
\toprule
       & \textbf{Fai.} & \textbf{Flu.} & \textbf{Inf.} & \textbf{Red.} & \textbf{Ave.} \\
\midrule
Zero           & 1.53 & \underline{1.83} & \textbf{1.94} & 1.44 & \underline{1.69} \\
GPT$_{\text{dir}}$ & \textbf{1.90} & 1.44 & 1.28 & \underline{1.47} & 1.52 \\
PbT            & \underline{1.87} & \textbf{1.94} & 1.14 & \textbf{2.00} & \textbf{1.74} \\
\hline
Real$_{\text{DSum}}$         & \underline{1.87} & \underline{1.47} & \underline{1.47} & \underline{1.47} & {1.57} \\
\bottomrule
\end{tabular}
\vspace{-2mm}
\end{table}

\paragraph{DialogSum}
We observe in Table \ref{tab:dialogsum} that in-context learning (ICL) delivers a substantial boost over the zero‐shot baseline, raising the R-L score from 19.6 to 28.7. Directly synthesizing source dialogues from target summaries with a large teacher model or a smaller student model yields performance slightly better or close to ICL. In addition, training on the DiaSynth~\citep{suresh2025diasynth} shows performance that is worse than the ICL results. 
This suggests that while applying strong LLM summarizers, directly generating synthetic data does not offer additional gains beyond what can be achieved by simply prompting the model with example summaries. 
The experiments emphasize the difficulty of using LLMs to synthesize data without actual paired source-target examples.
By contrast, our PbT methods deliver substantially greater gains than either ICL or direct synthesis. In the low‐resource setting, they raise R-L from 29.2 to 35.3, achieving 1.6 points below the oracle paired-data upper bound of 36.9. Moreover, this improvement holds consistently for both teacher models (GPT and LLAMA70B) and across different data scales.

\paragraph{SAMSum}
We conduct the same set of experiments on SAMSum. Table~\ref{tab:samsum_small} shows results for the small training setup (147-shot), and Appendix~\ref{sec:medium} for the medium (736-shot). Overall, SAMSum yields higher ROUGE scores in the zero-shot and ICL settings compared to DialogSum. Direct synthesis already provides gains over the ICL baselines. However, the PbT approach further improves R-L scores by 3\% absolute.
\subsection{Question Generation}
We evaluate PbT on the SQuAD dataset using the same training settings. 
Here, the source is a paragraph with its answer span, and the target is the corresponding question. 
The IR is defined as the sentence containing the answer, together with key facts and the answer span. 
Examples and prompts are provided in Appendix~\ref{sec:prompt}.

PbT outperforms both ICL and direct synthesis across all metrics, even surpassing oracle-paired training, mirroring the trend on CNNDM. 
Although zero-shot LLAMA8B achieves the best downstream QA scores, it lags far behind in BLEU/ROUGE and generates much longer questions than ground-truth. 
Direct synthesis recovers some lexical overlap (+2 ROUGE over the zero-shot baseline) but still trails IR-based PbT by 4 ROUGE-L and 2 BLEU, confirming the value of explicit IRs in question generation.



\begin{figure}[t]
  \centering
  \includegraphics[width=0.95\linewidth]{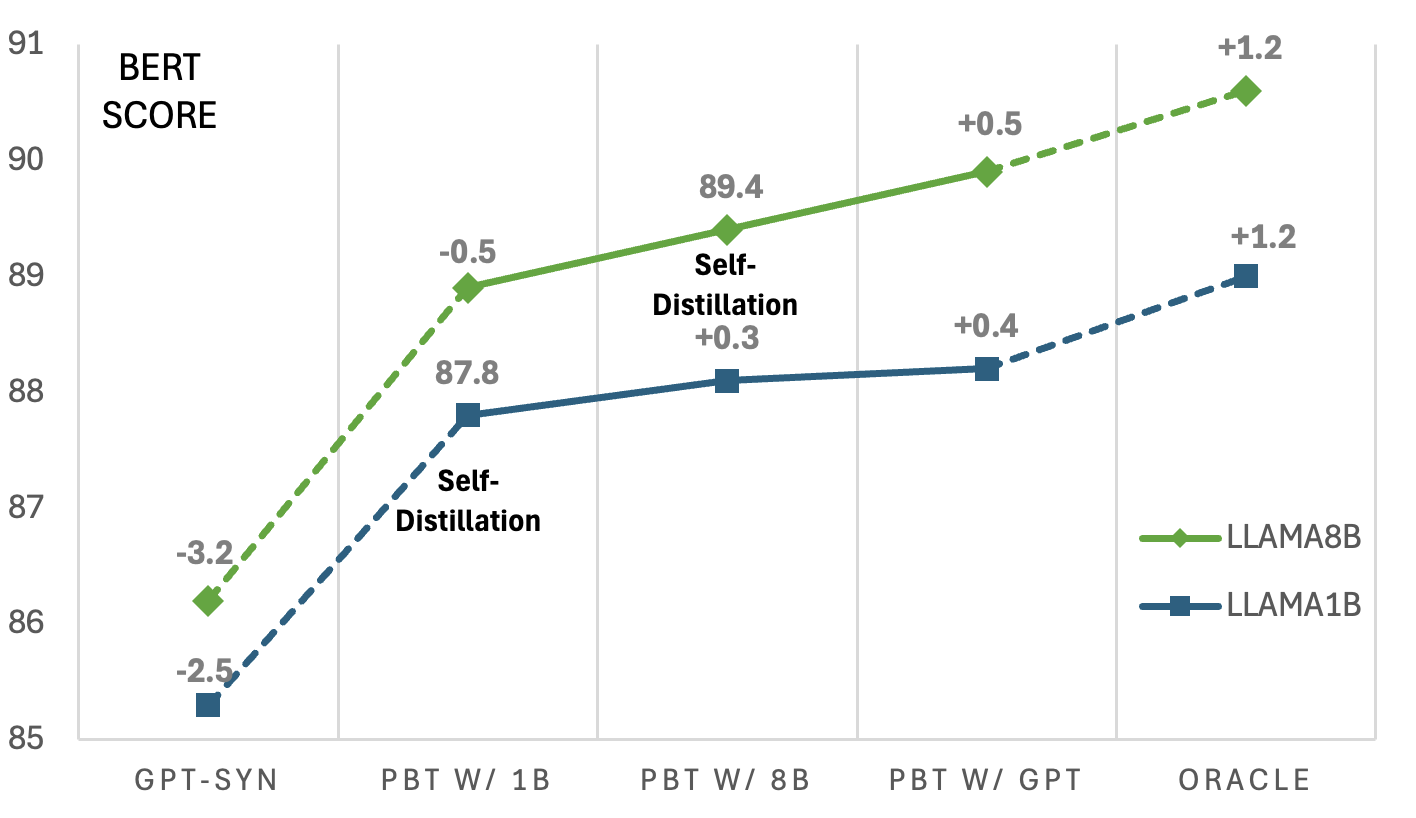}
  \caption{Direct synthesis from GPT, PbT w/ various teacher models, and real data comparison on XSum with LLAMA8B and LLAMA1B student models.}
  \label{fig:abl}
  \vspace{-2mm}
\end{figure}

\begin{table*}[h]
\centering
\caption{Teacher-side cost on XSum (LLAMA8B), measured on A100 GPUs. PbT requires only IR-level annotation ($m_x+m_y$), substantially reducing teacher effort while improving G-Eval scores.}
\label{tab:teacher_cost}
\begin{tabular}{lcccc}
\toprule
\textbf{Pipeline} & \textbf{Teacher-side Cost (min)} & \textbf{Rel. Cost} & \textbf{Saving (min)} & \textbf{G-Eval (Ave.)} \\
\midrule
Direct (full doc) & 224  & $1 \times$   & 0   & 2.75 \\
Direct w/ BoN-5   & 1120 & $5 \times$   & 0   & 2.88 \\
PbT (IR: $m_x+m_y$)   & 83 (22+61) & $\bm{0.37 \times}$ & 141 & 3.22 \\
PbT w/ BoN-5      & 415  & $1.85 \times$ & 705 & \textbf{3.43} \\
\bottomrule
\end{tabular}
\end{table*}

\begin{table}[t]
\label{tab:sub}
\centering
\caption{Different IR settings for PbT on XSum dataset. Teacher model: LLAMA8B, Student model: LLAMA1B}
\label{tab:abl_ir}
\begin{tabular}{@{}lcccc@{}}
\toprule
   & \textbf{Fact.} & \textbf{Coh.} & \textbf{Cont.} & \textbf{B-S}  \\
\midrule
PbT(IR$_\text{section}$) & \textbf{2.51} & \textbf{3.30} & \textbf{3.29} &  \textbf{88.1} \\
PbT(IR$_\text{hier}$) & 2.33 & 2.84  & 3.10 &  87.6
\\
PbT(IR$_\text{CoT}$)     & 2.43             & 2.95             & 3.12             &    87.9          \\
\bottomrule
\end{tabular}

\end{table}
\subsection{Cross Dataset Experiments}
We performed unpaired training combining SwitchBoard dialogues with DialogSum summaries, since SwitchBoard contains only raw conversations without summary annotations. 
We compare PbT against three baselines: zero-shot pretrained model, GPT-based direct synthesis, and a model trained on true DialogSum pairs and applied directly to SwitchBoard. Six human evaluators assessed the summaries in four dimensions--informativeness, faithfulness, fluency, and redundancy-- using a 0-2 scale, following the evaluation protocol from~\citep{lu2025mutual, xie2024few}.

Table~\ref{tab:swb} presents human evaluation on SwitchBoard datset. In faithfulness, all fine-tuned methods score similarly (1.87–1.90), outperforming the zero-shot model (1.53). PbT achieves the highest fluency (1.94), likely due to its training data being more closely aligned with SwitchBoard than the GPT${_\text{dir}}$ outputs or the Real${_\text{DSum}}$ pairs.
For informativeness and redundancy, we measure each model’s adherence to DialogSum’s concise two-to-three-sentence style. Summarizing long SwitchBoard dialogs naturally lowers informativeness (1.14) and reduces redundancy (2.00). 
PbT produces the most compact, on-target summaries, whereas other methods yield longer, more redundant outputs (1.44–1.47).
Overall, PbT ranks among the top performers on three of the four metrics, demonstrating robust, on-style summarization.  
Example summaries are provided in the Appendix~\ref{sec:examples}.


\subsection{Ablation Study}

\paragraph{Teacher Analysis}
Figure \ref{fig:abl} tests PbT with 8B and 1B students across six teacher–student pairings. In every setting, PbT surpasses the GPT-synthesis baseline. Even with the small 1B teacher guiding an 8B student, BERTScore improves by 2.7 points. For the 1B student, upgrading the teacher from 8B to GPT yields only a 0.1-point gain, indicating that oversized teachers offer limited additional benefit. These findings show that PbT remains effective even with modest teacher models, highlighting the framework’s robustness to teacher capacity.

\paragraph{Efficiency Analysis}
Table~\ref{tab:teacher_cost} compares teacher-side cost and quality across different pipelines on XSum-small. Direct synthesis requires 224 minutes of full-document annotation and achieves a G-Eval score of 2.75. Adding BoN-5 increases the cost fivefold (1120 minutes) while providing only a marginal gain (2.88). In contrast, PbT reduces annotation time to just 83 minutes by operating at the IR level ($m_x+m_y$), achieving a higher score of 3.22. When combined with BoN-5, PbT incurs 415 minutes and reaches the best quality of 3.43. These results demonstrate that PbT reduces teacher effort while improving summary quality.

\paragraph{IR types Analysis}
Under the LLAMA8B/1B setting, we compared three IR encoding methods—sectioned IR (our default), hierarchical IR, and chain-of-thought IR (Table~\ref{tab:abl_ir}). We used GPT-4o-mini to assess factual consistency, coherence, and content coverage of the generated documents. Results show that sectioned IR provides the highest overall quality, while all three variants remain the summarization scores within a half-point range in B-S, indicating stable performance across designs.

\section{Conclusion}
Paired by the Teacher (PbT) addresses the shortage of aligned input–output pairs in natural language generation by combining teacher-annotated intermediate representations with student-generated sources. This design converts unpaired corpora into usable supervision without human labeling, while shifting most text generation from the teacher to the student.
Across five benchmarks covering document summarization, dialogue summarization, and question generation, PbT consistently surpasses teacher- and student-only baselines. An 8B student trained with PbT data comes within 1.2 ROUGE-L points of human-labeled systems, closing 82\% of the gap over the strongest synthetic baseline at only one-third the annotation cost of direct synthesis. 
On the unlabeled SwitchBoard corpus, PbT further shows better style adherence and fluency than competing approaches. This advantage comes from PbT’s in-domain alignment: a student trained on real inputs generates sources close to the true distribution, avoiding the mismatch that limits direct synthesis.
PbT thus offers a fast, cost-effective way to train compact models for specialized or privacy-sensitive applications, even when labeled input–output pairs are scarce.


\section*{Limitations}
While synthesizing data, we train the student to reconstruct the source text in an open-ended manner. PbT generally introduces fewer hallucinations than directly synthesizing from the target. However, because the IR inevitably omits some details, the generated sources and summaries still require careful human review in high-fidelity applications.
All current experiments use English corpora only. We have not yet studied other languages or cross-lingual scenarios such as summarizing documents or dialogues written in a different language.


\section*{Ethical Considerations}
As we extend the frontiers of NLP, we remain attentive to the ethical dimensions of our work. Generating synthetic data with large‐scale models reduces our reliance on vast real‐world datasets that could include personal or sensitive information. To uphold privacy and meet regulatory standards, we draw exclusively on openly available corpora under permissive terms (CC BY-NC-ND 4.0 and CC BY-NC-SA 4.0) and enforce stringent anonymization measures.
When PbT is applied to copyrighted web or news material, the synthetic sources can echo portions of the originals. Practitioners should verify redistribution rights or ensure that outputs are sufficiently transformed. For conversational data, intermediate representations may reveal private entities or sensitive details, so names should be anonymized before teacher annotation. Finally, because PbT makes it easier to create large paired datasets, the same pipeline could be exploited to mass-produce malicious content. To limit misuse, we will release the code under a research-only license.


\bibliography{custom}

\appendix

\section{Appendix}
\label{sec:appendix}

%

\subsection{Training Settings}
\label{sec:hyp}
Our experiments mirror the setup in  ~\citep{lu2025mutual} for both data synthesis and summarization. We train with mini-batches of 10 and linearly ramp the learning rate up to $2.0 \times 10^{-4}$  over the first 50 batches. 
A ReduceLROnPlateau scheduler (patience = 5, factor = 0.7) handles learning-rate adjustments, and we invoke early stopping if the loss plateaus for 100 steps. 
Since the paired data are not available, we also generate the source data from the validation summaries to form the synthetic validation set for learning rate adjustment and model selection for summarization and question generation. 

\subsection{Medium Resource Experiments}
\label{sec:medium}
We include the medium resource (5\% dialogues and 1,000-shot summaries) experiments of XSum (Table~\ref{tab:xsum_medium}), SAMSum (Table~\ref{tab:samsum_medium}), and CNNDM (Table~\ref{tab:cnndm_medium}). Consistent improvement compared to the ICL and direct synthesis baselines can be observed, as in the small resource experiments.

\begin{table}[ht]
\centering
\caption{\textbf{XSum} (Medium, 1,000-shot unpaired documents): Comparison of methods on medium‑source setting}
\label{tab:xsum_medium}
\begin{tabular}{@{}lccc@{}}
\toprule
Synthesis model                     & R-2   & R-L   & BERTScore \\
\midrule
Zero shot                           &  5.3  & 12.5  & 86.2      \\
ICL (summary only)                  &  7.9  & 18.2  & 87.6      \\
\midrule
\multicolumn{4}{l}{\textbf{Unpaired Sources}} \\
\midrule
GPT$_{\text{direct}}$ ($\mathcal{T}$)        &  5.3  & 16.6  & 86.3      \\
LLAMA8B$_{\text{direct}}$ ($\mathcal{S}$)    &  6.6  & 18.6  & 87.0      \\
PbT$_\text{8B,\;w/\;GPT}$                   & \textbf{15.7} & \textbf{29.0} & \textbf{89.6} \\
\midrule
\multicolumn{4}{l}{\textbf{Oracle Paired}} \\
\midrule
Real Data                          & 18.9  & 32.7  & 90.6      \\
\bottomrule
\end{tabular}
\end{table}

\begin{table}[ht]
\centering
\caption{\textbf{SAMSum (736-shot unpaired dialogues)}: Comparison of methods on medium‑source setting.}
\label{tab:samsum_medium}
\begin{tabular}{@{}lccc@{}}
\toprule
Synthesis model            & R-2   & R-L   & B-S \\
\midrule
Zero‑shot                  & 12.9  & 25.1  & 88.3 \\
ICL (summary only)         & 18.5  & 33.8  & 90.4 \\
\midrule
\multicolumn{4}{l}{\textbf{Unpaired Sources}} \\
\midrule
GPT$_\text{direct}$              & 23.7  & 38.3  & 91.0 \\
LLAMA8B$_\text{direct}$          & 21.5  & 36.5  & 90.8 \\
DiaSynth                  & 16.0  & 29.8  & 89.5 \\
PbT$_\text{8B, w/ GPT}$       & \textbf{25.9} & \textbf{41.6} & \textbf{91.7} \\
\midrule
Oracle Paired              & 28.0  & 43.3  & 92.0 \\
\bottomrule
\end{tabular}
\end{table}

\begin{table}[ht]
\centering
\caption{\textbf{CNNDM (Medium, 1,000-shot unpaired documents)}: Comparison of methods on medium‑source setting.}
\label{tab:cnndm_medium}
\begin{tabular}{@{}lccc@{}}
\toprule
Synthesis model            & R-2   & R-L   & B-S \\
\midrule
Zero‑shot                  & 13.1  & 21.0  & 86.8 \\
ICL (summary‑only)         & 15.2    & 24.7    & 87.5   \\
\midrule
\multicolumn{4}{l}{\textbf{Unpaired Sources}} \\
\midrule
GPT$_\text{direct}$              & 16.6  & 23.6  & 86.6 \\
LLAMA8B$_\text{direct}$          & 17.3  & 25.1  & 86.9 \\
PbT$_\text{8B, w/ GPT}$       & \textbf{17.3} & \textbf{27.3} & \textbf{87.6} \\
\midrule
Oracle Paired              & \textbf{17.5} & 26.2 & 87.2 \\
\bottomrule
\end{tabular}
\end{table}

\subsection{Prompts for Intermediate Representation Extraction}
\label{sec:prompt}
We provide all the prompts for IR extraction from both sources and targets in Table~\ref{tab:ir_source} and Table~\ref{tab:ir_target}, including dialogue summarization, document summarization, and question generation. Also, the prompts for synthesizing source data from IR are shown in Table~\ref{tab:ir_syn}.

\subsection{Human Evaluation Details}
We recruited six graduate students to evaluate our dialogue summaries, with all participants providing informed consent for their contributions to be reported. Following the evaluation protocols established by  \citep{lu2025mutual, xie2024few}, annotators assessed each summary along four dimensions: faithfulness, fluency, informativeness, and redundancy. The guidelines for evaluation are outlined in Table~\ref{tab:human_eval_instruct}.

\subsection{Summarization and Question Generation Examples}
\label{sec:examples}

In this section, we show the source and generated target examples for each of the three tasks: dialogue summarization, document summarization, and question generation. The results are in Table~\ref{tab:summ_example_D},
Table~\ref{tab:summ_example_X}, and Table~\ref{tab:summ_example_S}. We also show the SwitchBoard dialogues summarization results with the target summaries examples in Table~\ref{tab:switchboard_input} and Table~\ref{tab:switchboard_output}.

\subsection{Document Generation Examples}
\label{sec:examples}
In this section, we show the real document and synthetic examples of XSum in Table~\ref{tab:syn_example_X} with IR-based PbT generation and directed synthesis baseline from the teacher.
We found the direct synthesis baseline generates text like a polished news article: it relies heavily on third-person narration, employs a wide range of complex sentence structures, and features frequent shifts in subjects and verb tenses. In contrast, PbT-generated text is straightforward, declarative sentences that mirror the style of real XSum data, using simple syntax and consistent phrasing to convey information directly.

\begin{table*}[]
\centering
\begin{tabular}{p{0.9\textwidth}}
\toprule
\textbf{Dialogue IR:}
You are an expert at segmenting dialogues and writing concise summaries. 
Given a raw  dialogue, divide it into segments and write one sentence of summary for roughly every 2–3 dialogue turns.  
The number of segments should adjust to the dialogue length.\\

Guidelines \\
1. Order: Follow the original flow of the conversation. \\
2. Density: Aim for 2–3 turns → 1 summary sentence. \\
3. Clarity: Each summary starts with “Segment N:” and is a single, clear sentence. \\
4. Content: Capture who does what, key facts, and any notable reactions or tone.   \\
   – Mention speaker names when it aids clarity.  \\ 
   – Do not quote full lines. \\
5. Output only the segment summaries: no additional commentary. \\

Output Format \\
Segment 1: … \\
Segment 2: … \\
Segment 3: … 
\\ 
\midrule
\textbf{Document IR:} You are an expert in document segmentation and summarization. Given an entire document, divide it into 3–7 coherent sections (introduction, key topics, conclusion, etc.). If any section is under 80 words, merge it with the previous one. \\

For each section, output exactly: \\ 

Section {n} \\ 
Summary: <1–2 sentences (20–30 words) capturing who/what, where, main action/result>  \\
Entities: <comma-separated list of key people, places, organisations>  
\\
(blank line)
\\ \midrule
\textbf{Question IR:}
You are an expert information extractor.
You will be given a paragraph and an answer span. Produce only the following \\
1) A single masked sentence with the answer replaced by [BLANK]. \\
2) Write 3–5 bullet points (•) capturing related facts (entities, actions, dates, numbers). \\
Do NOT include or repeat any other text from the paragraph. \\

Output format: \\
<masked sentence> \\
• bullet 1 \\ 
• bullet 2 \\ 
• bullet 3 \\ 
\bottomrule
\end{tabular}

\caption{Prompts for IR extraction from source}
\label{tab:ir_source}
\end{table*}

\begin{table*}[]
\centering
\begin{tabular}{p{0.9\textwidth}}
\toprule
\textbf{Dialogue IR:} You are an expert in dialogue reconstruction. Given a global summary of a dialogue and the approximate number of original dialogue lines,  produce a plausible set of segment summaries that would let a reader rebuild the dialogue flow. \\
Guidelines:\\
1. Segment density: 1 summary sentence for every 2–3 dialogue lines.\\
2. Order \& flow: Follow the chronology implied by the global summary.\\
3. Speaker cues: Mention speaker names when helpful, but do not quote full lines. \\
4. Content: Capture key actions, questions, answers, emotions, and shifts. \\
5. Format: Output only the segment summaries—no extra commentary. \\
Format:\\
Segment 1: …\\
Segment 2: …\\
…\\
6. Output only the segment summaries — no extra commentary or restating the global summary.
\\
\midrule
\textbf{Document IR:} You are an expert news editor. Given a global summary of an article and the article's  approximate total length in words 
Produce a set of segment summaries that partition the article logically 
Guidelines:\\
1. Density: Aim for 1 segment summary per 80-120 words\\
2. Order \& chronology: Preserve the timeline of events as implied by the global summary.\\
3. Content: Each segment summary (1-2 sentences) must cover key facts, actors, actions, and transitions; avoid speculative or external information.\\
4. Entities line: After each summary, list the main named entities (people, organisations, locations) separated by commas.\\
5. Format strictly as:\\
Section 1\\
Summary: <concise sentence(s)>\\
Entities: <Entity A>, <Entity B>, …\\
Section 2 … \\
6. Output only the segment summaries — no extra commentary or restating the global summary.
\\
 \midrule

\textbf{Question IR:}
You are an expert content planner. \\
Given a Question and the Answer, produce only the following:\\
1) One masked sentence that answers the question, with the answer replaced by [BLANK]. \\
2) Exactly 3–5 bullet points that would help a writer expand the idea into a full paragraph. \\
Guidelines: \\
• The masked sentence must be grammatically correct and fully answer the question when [BLANK] is filled. \\
• Each bullet <= 20 words and begins with •. \\
• Bullets should cover key entities, events, dates, figures, causes, etc.\\
• Do NOT repeat or rephrase any part of the question text.\\
• Output only the masked sentence and bullets—no extra commentary.\\
\bottomrule
\end{tabular}
\caption{Prompts for IR extraction from target}
\label{tab:ir_target}
\end{table*}

\begin{table*}[]
\centering
\begin{tabular}{p{0.9\textwidth}}
\toprule
\textbf{Dialogue:} You are a dialogue-generation model. Please follow these rules:\\1. You will be given segment summaries.\\ 2. Based solely on these summaries, reconstruct the conversation. \\3. Each turn starts with a speaker's name exactly as provided, followed by a colon.\\
4. Target length: about \{m\} turns and \{N\} words.\\
5. Do not output the segment summaries again or invent content beyond what is implied.\\ 6. Output must include a single section titled '=== Dialogue Begins ===\\ 7. Ensure the dialogue reads naturally and fluently.\\
\midrule
\textbf{Document:} You are a professional news-style writer. Follow these rules: \\
1. You will receive segment summaries describing an article. \\
2. Rely only on the information in those summaries—do not add facts that are not implied.\\
3. Write a coherent, well-structured document in clear journalistic prose.\\
4. Organise the text into logical paragraphs; you may merge or split ideas as long as all summary details are covered.\\
5. Target length: around \{N\} words\\
6. Do not repeat the segment-summary bullets in the output.\\
7. Start your answer with exactly the line:  === Document Begins ===\\
8. End with a newline; no extra commentary after the article.\\

\midrule
\textbf{Question:} 
You are a professional writer. Follow these rules:\\1. You will receive 3-5 bullet points (IR) and the answer span.\\2. Rely only on the information in those bullet points—do not add facts that are not implied.\\3. Write ONE coherent paragraph that uses all the bullet facts and naturally includes the answer.\\4. Target length: around\{N\} words.\\
5. Do not repeat the bullets in the output.\\ 6. Start your answer with exactly the line:  === Paragraph Begins ===\\7. End with a newline; no extra commentary after the paragraph.\\
\bottomrule
\end{tabular}
\caption{Prompts for student IR inversion in both training and testing.}
\label{tab:ir_syn}
\end{table*}

\begin{table*}[]
\centering
\begin{tabular}{l}
\toprule
\textbf{Faithfulness}\\ \midrule
Whether the information presented in the summary is factually incorrect or unmentioned \\
according to the dialogue: \\
*0: lots of the information presented in the summary is factually incorrect or unmentioned; \\
*1: a small amount of the information presented in the summary is factually incorrect or unmentioned;\\
*2: no information presented in the summary is factually incorrect or unmentioned.\\
\midrule
\textbf{Fluency}\\ \midrule
Whether the sentences in the summary are ungrammatical or ill-formed: \\
*0: lots of the sentences in the summary are ungrammatical or ill-formed; \\
*1: a small amount of the sentences in the summary are ungrammatical or ill-formed;\\
*2: no sentence in the summary is ungrammatical or ill-formed.\\
\midrule
\textbf{Informativeness}\\ \midrule
Whether the critical information in the dialogue is missed in the summary: \\
*0: lots of the critical information in the dialogue is missed; \\
*1: a small amount of the critical information in the dialogue is missed;\\
*2: no critical information in the dialogue is missed.\\
\midrule
\textbf{Redundancy}\\ \midrule
 Whether the expressions of the summary can be simplified: \\
*0: lots of the expressions of the summary can be simplified; \\
*1: a small amount of the expressions of the summary can be simplified;\\
*2: no expression of the summary can be simplified.\\
\bottomrule
\end{tabular}
\caption{Human evaluation metrics for Faithfulness, Fluency, Infomativeness, and Redundancy, and their corresponding instructions}
\label{tab:human_eval_instruct}
\end{table*}

\begin{table*}[]
\centering
\begin{tabular}{p{0.9\textwidth}}
\toprule
\textbf{Input Dialogue}\\ \midrule 
Person1: What are you working on?\\Person2: I'm figuring out my budget.\\Person1: What budget?\\Person2: I'm making a shopping budget, so that I don't spend too much money.\\Person1: How much money can you spend?\\Person2: I can only spend \$ 300 a month.\\Person1: Why only \$ 300?\\Person2: I need to save the rest.\\Person1: For what?\\Person2: I need to pay my bills.\\Person1: Your budget is a good idea.\\Person2: I know. It's going to save me a lot of money, I hope. \\
\midrule
\textbf{Ground Truth Summary} \\ \midrule
Person2 is making a shopping budget to save the rest of the money. Person1 thinks it's a good idea.\\
\midrule
\textbf{Generated Summary (Zero Shot)} \\ \midrule
Person 1 asks Person 2 what they are working on, and Person 2 responds that they are creating \\a shopping budget to avoid overspending. \\Person 1 asks how much they can spend, and Person 2 says \$300 per month. \\Person 1 asks why they can only spend \$300, \\ and Person 2 explains that they need to save the rest to pay their bills. \\Person 1 thinks the budget is a good idea, and Person 2 hopes it will help them save money.\\
\midrule

\textbf{Generated Summary (In-context Learning)} \\
\midrule  
Person1 asks Person2 what they are working on, and Person2 explains that they are creating \\ a budget to control their spending. \\Person2 sets a limit of \$300 per month and plans to save the rest for paying bills.\\Person1 praises Person2's budgeting idea, which Person2 hopes will help them save money.\\
\midrule




\textbf{Generated Summary (Direct Synthesis, GPT)} 
\\ \midrule Mary is making a shopping budget so that she doesn't spend too much money. \\Person1 asks her how much money she can spend and why only \$ 300. \\Mary says she needs to save the rest to pay her bills.\\
\midrule
\textbf{Generated Summary (PbT, Ours)} \\ \midrule
Person2 is making a budget to save money. Person1 thinks it's a good idea.\\
\midrule
\textbf{Generated Summary (Real Data)} \\ \midrule
Person2 is making a shopping budget so that Person2 doesn't spend too much money.\\
\bottomrule
\end{tabular}
\caption{DialogSum Summarization results from PbT and baselines.}
\label{tab:summ_example_D}
\end{table*}

\begin{table*}[]
\centering
\begin{tabular}{p{0.9\textwidth}}
\toprule
\textbf{Input document}\\ \midrule 
Their recent run is in contrast to the way they started the season with six victories from their first seven games.Since the end of March, the Vikings' only wins have been in the Challenge Cup against lower-league sides."We've got the personnel and we've got the people to spark us back into life," Betts told BBC Radio Merseyside."When we get rolling again I'm sure, or I'm positive, that we can really turn this year around for ourselves."Widnes lie eighth after their poor run and face Salford at Magic Weekend on Saturday."The players are hurting and we've got to win," added England assistant coach Betts. "Whether it be pretty, whether it be the last second of the game, we've got to win."It's alright being a good group and it's alright talking about team spirit and being in it together, but not winning really hurts."We've got to find that win and we've got to find it some way and that's within the team and it's in those 80 minutes out there."\\
\midrule
\textbf{Ground Truth Summary} \\ \midrule
Widnes head coach Denis Betts is confident his squad will find their form after a seven-match losing streak in Super League.
\\
\midrule
\textbf{Generated Summary (Zero Shot)} \\ 
\midrule
Widnes Vikings, an English rugby league team, have struggled since the end of March,  winning only against lower-league teams in the Challenge Cup. Despite having the talent and team spirit, they are currently in 8th place and need to turn their season around. The team's assistant coach, Betts, emphasizes the importance of winning,  stating that it's not just about being a good team, but about getting results. They face Salford at Magic Weekend on Saturday and are determined to find a way to win.\\

\midrule

\textbf{Generated Summary (In-context Learning)} \\
\midrule  
Widnes Vikings are struggling in the season, having only won in the Challenge Cup against lower-league teams  since the end of March. Assistant coach Betts is confident that the team can turn their season around, citing their personnel and team spirit as reasons for optimism. They face Salford at Magic Weekend on Saturday. \\

\midrule 
\textbf{Generated Summary (Direct Synthesis, GPT)} 
\\ \midrule 
Their run of six wins from their first seven games is the opposite of the way  they started the season with six from their first seven.
\\
\midrule
\textbf{Generated Summary (PbT, Ours)} \\ \midrule
Widnes Vikings coach Denis Betts says his team are "hurting" after a run of 10 Super League defeats in a row.\\
\midrule
\textbf{Generated Summary (Real Data)} \\ \midrule
Widnes Vikings head coach Denis Betts says he is confident his side can turn their season around after a run of 10 Super League defeats.
 \\
\bottomrule
\end{tabular}
\caption{XSum Summarization results from PbT and baselines.}
\label{tab:summ_example_X}
\end{table*}

\begin{table*}[]
\centering
\begin{tabular}{p{0.9\textwidth}}
\toprule
\textbf{Input paragraph and Answer Span} \\ \midrule 
\textbf{Paragraph:} In October 2010, the open-access scientific journal PLoS Pathogens published a paper by a multinational team who undertook a new investigation into the role of Yersinia pestis in the Black Death following the disputed identification by Drancourt and Raoult in 1998. They assessed the presence of DNA/RNA with Polymerase Chain Reaction (PCR) techniques for Y.  pestis \textbf{$\textbf{[}$from the tooth sockets in human skeletons$\textbf{]}$} from mass graves in northern, central and southern Europe that were associated archaeologically with the Black Death and subsequent resurgences.  The authors concluded that this new research, together with prior analyses from the south of France and Germany, ". . . ends the debate about the etiology of the Black Death, and unambiguously demonstrates that Y.  pestis was the causative agent of the epidemic plague that devastated Europe during the Middle Ages". \\
\textbf{Answer:} from the tooth sockets in human skeletons
\\
\midrule
\textbf{Ground Truth Question} \\ \midrule
Where did scientists find their Y. pestis sample? 
\\
\midrule
\textbf{Generated Question (Zero Shot)} \\ \midrule
Where did a multinational team use Polymerase Chain Reaction (PCR) techniques to \\ assess the presence of Y. pestis DNA/RNA?
\\
\midrule
\textbf{Generated Question (In-context Learning)} \\
\midrule  
Where did a multinational team assess the presence of DNA/RNA with \\Polymerase Chain Reaction (PCR) techniques?
\\
\midrule 
\textbf{Generated Question (Direct Synthesis, GPT)} 
\\ \midrule 
Where did the new investigation of Yersinia pestis take place?
\\
\midrule
\textbf{Generated Question (PbT, Ours)} \\ \midrule
Where did the researchers assess the presence of DNA/RNA with PCR techniques for Y. pestis?
\\
\midrule
\textbf{Generated Question (Real Data)} \\ \midrule
Where did the researchers look for Yersinia pestis? \\
\bottomrule
\end{tabular}
\caption{SQuAD Question Generation results from PbT and baselines.}
\label{tab:summ_example_S}
\end{table*}

\begin{table*}[t]
\centering
\begin{tabular}{p{0.9\textwidth}}
\toprule
\textbf{Input Dialogue (SwitchBoard, Unpaired)} \\
\midrule
A: Hi, my name is Donna, and  Im  calling,  calling  from Plano, Texas.\\B: Im Jay in Dallas.\\A: Oh,  okay,  um, I,  um,  find exercise kind of a task,  I guess Id say. I go to aerobics  two days a week, but  I also have  three kids at home so  they keep me pretty busy.\\B: That keeps you plenty of exercise      .\\A: Yeah, that gives me plenty of exercise, but  I enjoy,  um,   walking outside and  bicycling once in a while with the family. Excuse me  . Thats fine. And yourself.\\B: Well,   I,   uh,  I   have, - I walked and run  ev-,   um,  almost every  morning. Ive sort of stopped here with the weather change, but,   uh,  its about a thirty minute exercise every morning.\\A: Uh-huh, every morning. Well,  my husband swims every day at lunch, which he loves doing --\\B: Oh,  that would, - thats fantastic exercise.\\A: -- yeah, he does it every day like a ritual. Once in a while or if he has a business meeting or if  he,   um,  someone   invites him  to lunch,  hell,  hell  not go, but  he goes,  um,   every,  every   day, and then   he ru-,  hes a runner,  too. He runs three mornings a week he gets up about five   and,  and   --\\B: Um.\\A: -- does  his,  his  deal, and  some days he doesnt, but  then maybe hell catch up on the weekend and go for a run.\\B: All right.\\A: And,   um,  we      . -\\B: Id rather get a bicycle, but   I have an ex-,  I had an exercise  bike in the house, but didnt seem to  use it enough, so  somebody sold it in a garage sale, and so,     uh  --\\A: Oh.\\B: --  but   Id like to get a bike and just  do,   uh,  do  biking outside.\\A: The,  the,   the  bicycling is nice especially with the weather around here, and  I know,  um,   uh,     we,  we    both  of,    my husband and I both   do it to just  keep the pounds off. Were getting up in age, I guess, in our late thirties so  we need to --\\B: Yeah.\\A: -- keep exercising just to be able to,  uh.  -\\B: It keeps going, because  Im fifty-five, and   it,    uh,  it  just gets worse. You still want to eat, and,     uh  --\\A: Oh,  I know.\\B: --  but   you just, - I think you feel better, too, if you get on some program like that, but,   uh.  -\\A: Yeah, I go to this aerobics class two mornings a week, and  I take my three year old with me. Theres a nursery there. And  its kind of another day out for her --\\B: Uh-huh.\\A: -- to,  um,  have little friends, and,   um,   I,  I  enjoy the socialization of it,  you know, getting to see the same women every --\\B: Right.\\A: -- morning, but   um,  there are some parts during that exercise that I think, I cannot do  another --\\B: Oh.\\A: -- second of this.\\B: I know    it,  it,   it,   some of it  is so tough --\\A: It is.\\B: --  and,   uh,  especially when somebodys,  uh,  planning lunch for you and they know  what,  whats  good for you, I think when we do it ourselves, we take the easy way sometimes.\\A: Thats true, thats true.\\
\bottomrule
\end{tabular}
\caption{Input dialogue from SwitchBoard used for summarization evaluation.}
\label{tab:switchboard_input}
\end{table*}

\begin{table*}[]
\centering
\begin{tabular}{p{0.9\textwidth}}
\toprule
\textbf{Example Summaries (DialogSum, Unpaired)} \\ \midrule
B sleeps through a movie and gets woken up by A. B thinks the movie is unrealistic while A thinks it was great. A doubts how B knows. B doesn't answer and suggests going to bed. \\ 
\midrule
\textbf{Generated Summary (Zero Shot)} \\ \midrule
The conversation is between Donna from Plano, Texas, and Jay from Dallas, Texas, who are participating in a phone survey about their exercise habits. Here's a summary of their conversation:  * Donna mentions that she finds exercise a task, but enjoys walking and bicycling with her family. * Jay says he used to walk and run every morning, but has stopped due to the weather change. * Donna's husband swims every day at lunch and runs three mornings a week. * Jay wants to get a bicycle to do outdoor biking, but his exercise bike was sold in a garage sale. * Both Donna and Jay mention that they exercise to keep the pounds off and stay healthy, especially as they get older. * Donna takes her three-year-old daughter to an aerobics class with a nursery, which provides socialization for her child. * Both Donna and Jay agree that some exercises can be tough, and it's easy to take the easy way out when planning meals. * They wrap up the conversation by thanking each other for participating in the survey.
\\ \midrule
\textbf{Generated Summary (In-context Learning)} \\ \midrule
Donna from Plano, Texas, and Jay from Dallas discuss their exercise routines. Donna mentions that she goes to aerobics classes two days a week and enjoys walking and bicycling with her family. Jay says he used to walk and run every morning but has stopped due to the weather change. Donna's husband swims every day at lunch and runs three mornings a week. They both agree that regular exercise is important, especially as they get older, to keep the pounds off and feel better.
\\ \midrule
\textbf{Generated Summary (Direct Synthesis, GPT)} \\ \midrule
Donna calls from Plano, Texas. Donna finds exercise a task. She goes to aerobics two days a week and has three kids at home so they give her plenty of exercise. Donna enjoys walking outside and bicycling once in a while with the family. B says B has walked and run almost every morning. B's husband swims every day at lunch which is fantastic exercise. B's husband is a runner and runs three mornings a week. B wants to get a bicycle but B had an exercise bike in the house which somebody sold in a garage sale. \\
\midrule
\textbf{Generated Summary (PbT, Ours)} \\ \midrule
Donna and Jay talk about their exercise habits. Donna goes to aerobics two times a week and enjoys walking and bicycling with her family. Jay runs every morning and wants to get a bicycle. \\
\midrule
\textbf{Generated Summary (DialogSum Real Data)} \\ \midrule
Donna from Plano, Texas calls Jay in Dallas. Donna has three kids at home so they keep her pretty busy. She enjoys walking outside and bicycling once in a while with the family. Her husband swims every day at lunch and he runs three mornings a week. They both do bicycling to keep the pounds off. Jay has an exercise bike in the house but didn't seem to use it enough. Jay likes to get a bike and just do biking outside. \\
\bottomrule
\end{tabular}
\caption{Summarization results from PbT and baselines on the SwitchBoard dialogue in Table~\ref{tab:switchboard_input}.}
\label{tab:switchboard_output}
\end{table*}

\begin{table*}[]
\centering
\begin{tabular}{p{0.9\textwidth}}
\toprule
\textbf{Target summary}\\ \midrule 
US actor Randy Quaid is due to appear in court in Vermont on Monday over a California vandalism case.\\
\midrule
\textbf{Real Document} \\
(A news-reporting style with a neutral tone, using precise descriptions of locations, times, and amounts) \\\midrule
Quaid, 65, and his wife, Evi, were taken into custody on Friday night by Vermont police while trying to enter the US from Canada. They fled to Canada in 2010 after facing charges of vandalism and squatting in a Santa Barbara property. But his request for refugee status, in which he claimed his life was in danger from organised crime, had been denied. Quaid was recently informed he would be sent back to the US to face criminal charges. His wife has dual US-Canadian citizenship. The Quaids are alleged to have caused more than \$5,000 (\pounds3,253) in damage to a property in Santa Barbara, which they claim belongs to them. Santa Barbara officials have previously said they would arrest Quaid if he returned to the US. At a hearing in 2010, he told a Canadian court he was being targeted by "Hollywood star whackers", and that eight of his fellow actors had been "murdered in odd, strange manners", including Heath Ledger and David Carradine. Quaid, the older brother of fellow actor Dennis, is best known for his roles in such films as Kingpin, National Lampoon's Vacation and The Last Detail. He and his wife Evi have been posting bizarre expletive-ridden videos on their YouTube channel, denouncing the US authorities' attempts to force them back.\\
\midrule
\textbf{Generated Documents (Direct Synthesis, GPT)}  \\

(Adopts a commentary-and-interpretation tone, inserting embellishments like “media attention” and “Fans and critics,” which clash with the original’s investigative news style.)\\ \midrule
US actor Randy Quaid is set to appear in a Vermont court on Monday as part of ongoing legal troubles stemming from a vandalism case in California. The 73-year-old actor, known for his roles in films such as "National Lampoon's Vacation" and "Independence Day," was arrested in 2020 after allegedly causing damage to a home in California. The incident sparked significant \textbf{media attention}, particularly due to Quaid's erratic behavior and public outbursts in recent years.  In a series of social media posts, Quaid has maintained his innocence, claiming he was targeted unfairly. His legal team has argued that the vandalism charges are exaggerated and stem from a personal dispute. The actor has previously faced several legal issues, including arrests related to trespassing and other charges.  Quaid's upcoming court appearance in Vermont relates to a separate matter, but his past incidents continue to cast a shadow over his career. \textbf{Fans and critics} alike have expressed concern over his well-being, with some calling for him to seek help for mental health issues.   As the court date approaches, many are watching closely to see how this high-profile case will unfold. Legal experts suggest that Quaid's history may affect the outcome of the proceedings. Regardless of the court's decision, the actor's reputation and future in Hollywood remain uncertain.\\
\bottomrule
\end{tabular}
\caption{XSum Document Synthesis examples results from PbT and baselines (Part 1).}

\label{tab:syn_example_X_a}
\end{table*}

\begin{table*}[t]
\centering
\begin{tabular}{p{0.9\textwidth}}
\toprule
\textbf{Generated Documents (PbT, Ours)} \\ 
(Direct stating tone, similar to the real XSum document.) \\
\midrule
Randy Quaid, best known for playing the evil brother Lyle in National Lampoon's Vacation, has also made headlines recently in California where his wife Evi is charged with looting two hotels while she was facing foreclosure. Quaid and his wife face charges over \$5m (\pounds3.15m) damages at several hotels they stayed in. According to Vermont officials, Randy Quaid would fly up to attend a court hearing if needed. The court date concerns a charge of vandalism by a tenant who had been locked out or excluded from a rented building. The allegations come under Vermont state law but were originally made in California. Vermont Attorney General William Sorrell said it did not matter where Mr Quaid lived, as he was being sued there because of events alleged to have happened there. Prosecutors allege the couple broke windows and doors at several homes including in Marin County and in San Francisco. Mr Quaid is also accused of leaving a threatening note saying "you're a victim", authorities added. The couple lost their home in Tiburon last year. In December, the sheriff's office took Mrs Quaid and her husband into custody and booked them at jail on suspicion of burglary and vandalism - the same charges prosecutors said they planned to pursue. Their children and grandchildren still live in the properties the couple are accused of damaging, according to media reports. They had moved to Canada to work out a possible deal allowing them to stay temporarily. But California lawyers say they failed to complete what they promised and return home.
 \\
\bottomrule
\end{tabular}
\caption{XSum Document Synthesis examples results from PbT and baselines (Part 2).}
\label{tab:syn_example_X}
\end{table*}

\end{document}